\documentclass[%
 reprint,
 amsmath,amssymb,
 aps, 
 pre,
 showkeys
]{revtex4-1}

\usepackage{graphicx}
\usepackage{dcolumn}
\usepackage{bm}

\usepackage{my-style}
\graphicspath{{figs/}}
\usepackage{algpseudocode, algorithm}
\usepackage{multirow}
\usepackage{color}

\newtheorem{theorem}{Theorem}

\allowdisplaybreaks[4]

\begin{document}

\preprint{APS/123-QED}

\title{Free Energy Evaluation Using Marginalized Annealed Importance Sampling}

\author{Muneki Yasuda}  
\email{muneki@yz.yamagata-u.ac.jp}
\affiliation{%
 Graduate School of Science and Engineering, Yamagata University, Japan.
}%
\author{Chako Takahashi}  
\affiliation{%
 Graduate School of Science and Engineering, Yamagata University, Japan.
}%


\begin{abstract}
The evaluation of the free energy of a stochastic model is considered a significant issue in various fields of physics and machine learning. 
However, the exact free energy evaluation is computationally infeasible because the free energy expression includes an intractable partition function. 
Annealed importance sampling (AIS) is a type of importance sampling based on the Markov chain Monte Carlo method that is similar to a simulated annealing 
and can effectively approximate the free energy. 
This study proposes an AIS-based approach, which is referred to as marginalized AIS (mAIS). 
The statistical efficiency of mAIS is investigated in detail based on theoretical and numerical perspectives.  
Based on the investigation, it is proved that mAIS is more effective than AIS under a certain condition.
\end{abstract}

\pacs{Valid PACS appear here}
\keywords{free energy evaluation, annealed importance sampling, Rao--Blackwellization, Ising model, restiricted Boltzmann machine}
\maketitle


\section{Introduction}
\label{sec:intro}
The evaluation of the free energy (i.e., the negative log of the partition function) of a stochastic model is considered to be an important issue in various fields of physics.
Moreover, the free energy plays an important role in machine learning models, e.g., Boltzmann machine~\cite{PBM1985}. 
Boltzmann machine and its variants, e.g., restricted Boltzmann machine (RBM)~\cite{RBM1986,CD2002} and deep Boltzmann machine (DBM)~\cite{DBM2009}, 
have been actively investigated in the fields of machine learning and physics~\cite{Roudi2009,decelle2021,chen2018,nomura2021,torlai2018,carleo2017,Gao2017}.
However, an exact free energy evaluation is computationally infeasible, because the expression for the free energy includes an intractable partition function. 

\textit{Annealed importance sampling} (AIS)~\cite{AIS2001} is a type of importance sampling based on the Markov chain Monte Carlo (MCMC) method that is similar to simulated annealing, 
and can effectively approximate the free energy~\cite{AIS2001,Salakhutdinov2008}.
In AIS, from a tractable initial distribution to the target distribution, a sequential sampling (or ancestral sampling)  is executed, 
in which the transitions between the distributions are performed using, for example, Gibbs sampling~\cite{Geman&Geman1984}.  
The AIS-based free energy evaluation is essentially the same as the method proposed by Jarzynski (the so-called Jarzynski equality)~\cite{Jarzynski1997}.  
Several researchers have addressed the development of AIS~\cite{RAISE2015,DS2015,carlson2016,mazzanti2020,krause2020,sekimoto2021}. 
Recently, the relationship between AIS and the normalizing flow proposed in the deep learning field is investigated~\cite{Caselle2022}. 

In AIS, the free energy is estimated as follows: 
first, we obtain the estimator of the partition function based on the sample average of \textit{the importance weights}, 
which are obtained during the sequential sampling process, and subsequently, 
the free energy estimator is obtained as the negative log of the obtained partition function estimator.
It is known that the partition function estimator is unbiased, whereas the free energy estimator is biased~\cite{RAISE2015}.
To evaluate the free energy, we propose an AIS-based approach that we refer to as \textit{marginalized AIS} (mAIS). 
The concept of mAIS is simple; mAIS corresponds to AIS in a marginalized model; i.e., mAIS can be regarded as a special case of AIS.
Therefore, the basic principle of mAIS is similar to AIS. 
Suppose that our target model represents a distribution in $n$ dimensional space. 
With regard to AIS, we perform a sampling procedure in the $n$ dimensional space to evaluate the free energy. 
However, in mAIS, the dimensional space, where the sampling procedure is performed, is smaller because the dimension of the model is reduced through marginalization. 
Intuitively, mAIS seems to be more effective than AIS considering the aforementioned statement.  
This intuition is valid under a certain condition. 
Under the condition, the following two statements can be proved (see section \ref{sec:AIS_vs_mAIS}): 
(1) the partition function estimator obtained from mAIS is more accurate than that obtained from AIS,  
and (2) the bias of the free energy estimator obtained from mAIS is lower than that of the estimator obtained from AIS.
However, as discussed in section \ref{sec:AIS_vs_mAIS}, the condition assumed in the aforementioned statements limits the range in which the effectiveness of mAIS is assured. 
As discussed in section \ref{sec:application_to_BT_MRF}, 
this condition can be satisfied with regard to the use of AIS and mAIS on Markov random fields (MRFs) defined on bipartite graphs. 
Moreover, a layer-wised marginal operation can be performed in bipartite-type MRFs. 
Bipartite-type MRFs include important applications, e.g., RBM and DBM. 

The remainder of this paper is organized as follows: 
Section \ref{sec:AIS} explains the AIS-based free energy evaluation. 
Section \ref{sec:proposed_sec} introduces mAIS 
Section \ref{sec:mAIS} details mAIS. 
Section \ref{sec:AIS_vs_mAIS} describes the theoretical analysis of mAIS, 
and section \ref{sec:application_to_BT_MRF} discusses the application of mAIS to bipartite-type MRFs. 
Section \ref{sec:experiment} numerically demonstrates the validity of mAIS. 
Section \ref{sec:summary} summarizes the study, along with discussions.

\section{Free Energy Evaluation using Annealed Importance Sampling}
\label{sec:AIS}

Consider a distribution with $n$ random variables, $\bm{x} := \{x_i \in \mcal{X}_i \mid i \in U := \{1,2,\ldots, n\}\}$, 
as
\begin{align}
P_{\mrm{model}}(\bm{x}):=\frac{1}{Z} \exp\big( - E(\bm{x})\big),
\label{eqn:model}
\end{align}
where $\mcal{X}_i$ is the continuous or discrete sample space of $x_i$ and $E(\bm{x})$ is the energy function (or the Hamiltonian); 
$Z$ is the partition function and is defined as
\begin{align}
Z: = \sum_{\bm{x}} \exp\big( - E(\bm{x})\big),
\label{eqn:Z}
\end{align}
where $\sum_{\bm{x}} := \sum_{x_1 \in \mcal{X}_1}\sum_{x_2 \in \mcal{X}_2} \cdots \sum_{x_n \in \mcal{X}_n}$ denotes the multiple summation over all realizations of $\bm{x}$.
When $\mcal{X}_i$ exhibits a continuous space, $\sum_{x_i}$ is replaced by the integration over $\mcal{X}_i$.

This study evaluates free energy $F:= - \ln Z$. 
The evaluation of the free energy is infeasible because it requires the evaluation of the intractable partition function.
AIS is a type of importance sampling based on the MCMC method that is similar to simulated annealing, 
and can evaluate the free energy~\cite{AIS2001,Salakhutdinov2008}. 
This free energy evaluation method is essentially the same as the method proposed by Jarzynski~\cite{Jarzynski1997}. 
The AIS-based free energy evaluation is briefly explained in the following.

First, we design a sequence of distributions as
\begin{align}
\{P_k(\bm{x}) \mid k = 0,1,\ldots, K\},
\label{eqn:sequence_distribution}
\end{align}
where $P_K(\bm{x}) = P_{\mrm{model}}(\bm{x})$, and each $P_k(\bm{x})$ is expressed as
\begin{align}
P_k(\bm{x}):= \frac{1}{Z_k} \exp\big( - E_k(\bm{x})\big),
\label{eqn:P_k(x)}
\end{align}
where $Z_k$ is the partition function of the $k$th distribution.  
Because $P_K(\bm{x}) = P_{\mrm{model}}(\bm{x})$, $E_K(\bm{x}) = E(\bm{x})$ and $Z_K = Z$ are assumed. 
In this sequence, the initial distribution $P_0(\bm{x})$ is set to a tractable distribution (e.g., a uniform distribution).
For example, the sequence expressed as 
\begin{align}
P_k(\bm{x}) \propto P_0(\bm{x})^{1 -\beta_k}P_{\mrm{model}}(\bm{x})^{\beta_k},
\label{eqn:P_k(x)_beta_k}
\end{align}
for $0= \beta_0 < \beta_1 < \cdots < \beta_K = 1$, is frequently used, where $\beta_k$ corresponds to the annealing temperature.
The $k$th distribution expressed by equation (\ref{eqn:P_k(x)_beta_k}) corresponds to equation (\ref{eqn:P_k(x)}) with  
$E_k(\bm{x}) = (1 - \beta_k) \ln  P_0(\bm{x}) + \beta_k E(\bm{x})$. 
However, we pursue the following arguments without specifying the form of $E_k(\bm{x})$. 

On the $k$th distribution, a transition probability $T_k(\bm{x}' \mid \bm{x})$, which satisfies the condition
\begin{align}
P_k(\bm{x}') = \sum_{\bm{x}} T_k(\bm{x}' \mid \bm{x}) P_k(\bm{x}),
\label{eqn:balance_condition}
\end{align}
is defined. Using the transition probability, the annealing process from the initial state $\bm{x}^{(1)}$ to the final state $\bm{x}^{(K)}$ is defined as
\begin{align}
T(\bm{X}):= \Big( \prod_{k=1}^{K-1} T_k(\bm{x}^{(k+1)} \mid \bm{x}^{(k)})\Big)P_0(\bm{x}^{(1)}),
\label{eqn:annealing_process}
\end{align}
where $\bm{X}:= \{\bm{x}^{(k)} \mid k = 1,2,\ldots, K\}$. $T(\bm{X})$ denotes the joint distribution over $\bm{X}$.
For $T(\bm{X})$, the corresponding ``reverse'' process,
\begin{align}
\tilde{T}(\bm{X}):= \Big( \prod_{k=1}^{K-1} \tilde{T}_k(\bm{x}^{(k)} \mid \bm{x}^{(k+1)})\Big)P_K(\bm{x}^{(K)}),
\label{eqn:reverse_process}
\end{align} 
is defined, where 
\begin{align}
\tilde{T}_k(\bm{x} \mid \bm{x}'):=\frac{T_k(\bm{x}' \mid \bm{x}) P_k(\bm{x})}{P_k(\bm{x}')}
\label{eqn:reverse_transition}
\end{align}
is the reverse transition. 
$T(\bm{X})$ expresses the transition process from $\bm{x}^{(1)}$ to $\bm{x}^{(K)}$;  
while, its reverse process $\tilde{T}(\bm{X})$ expresses the reverse transition from $\bm{x}^{(K)}$ to $\bm{x}^{(1)}$.
Because the normalization condition, $\sum_{\bm{x}}\tilde{T}_k(\bm{x} \mid \bm{x}') = 1$, is ensured from the condition in  equation (\ref{eqn:balance_condition}), 
$\tilde{T}(\bm{X})$ can be considered as the joint distribution over $\bm{X}$;  it satisfies 
\begin{align*}
\sum_{\bm{x}^{(1)}}\sum_{\bm{x}^{(2)}}\cdots \sum_{\bm{x}^{(K-1)}}\tilde{T}(\bm{X}) = P_K(\bm{x}^{(K)}).
\end{align*}
AIS is regarded as the importance sampling, in which $\tilde{T}(\bm{X})$ and $T(\bm{X})$ are considered the target  
and corresponding proposal distributions, respectively. 
Using equations (\ref{eqn:annealing_process}), (\ref{eqn:reverse_process}), and (\ref{eqn:reverse_transition}), 
the ratio between the target and proposal distributions, i.e., the (normalized) importance weight, is obtained as 
\begin{align}
\frac{\tilde{T}(\bm{X})}{T(\bm{X})}= \frac{Z_0}{Z_K}W(\bm{X}),
\label{eqn:AIS_IW}
\end{align}
where
\begin{align}
W(\bm{X})&:=\prod_{k=1}^Kw_k(\bm{x}^{(k)}),
\label{eqn:AIS_IW_w}\\
w_k(\bm{x})&:= \exp\big( - E_k(\bm{x})  + E_{k-1}(\bm{x}) \big).
\label{eqn:AIS_IW_wk}
\end{align}
Here, $Z_0$ is the partition function of $P_0(\bm{x})$ (i.e., a tractable distribution) and $Z_K = Z$ is the true partition function of the objective distribution.  
From the normalization condition of $\tilde{T}(\bm{X})$, the relation,
\begin{align*}
1 =  \sum_{\bm{X}} \frac{\tilde{T}(\bm{X})}{T(\bm{X})}T(\bm{X}),
\end{align*}
is obtained. Substituting equation (\ref{eqn:AIS_IW}) into this relation yields 
\begin{align}
Z_K = Z_0 \sum_{\bm{X}} W(\bm{X})T(\bm{X}),
\label{eqn:Zk_AIS}
\end{align}
where $\sum_{\bm{X}} := \sum_{\bm{x}^{(1)}}\sum_{\bm{x}^{(2)}}\cdots \sum_{\bm{x}^{(K)}}$ denotes the multiple summation over all realizations of $\bm{X}$.

Sampling from the proposal distribution can be performed using ancestral sampling, from $\bm{x}^{(1)}$ to $\bm{x}^{(K)}$, on  $T(\bm{X})$,  
i.e., the sequence of the sample points, $\mbf{X}:=\{\mbf{x}^{(1)}, \mbf{x}^{(2)}, \ldots, \mbf{x}^{(K)}\}$, is generated by the following sampling process:
\begin{align}
\begin{split}
\mbf{x}^{(1)}&\leftarrow P_0(\bm{x}),\\
\mbf{x}^{(k)}&\leftarrow T_{k-1}(\bm{x} \mid \mbf{x}^{(k-1)})\>\> (k = 2,3,\ldots,K).
\end{split}
\label{eqn:AIS_sampling}
\end{align}
Using $N$ independent sample sequences $S_{T} :=\{\mbf{X}_{\mu} \mid \mu = 1,2,\ldots,N\}$ obtained when this ancestral sampling process conducted $N$ times, 
equation (\ref{eqn:Zk_AIS}) is approximated as
\begin{align}
Z \approx Z_{\mrm{AIS}}(S_T):=\frac{Z_0}{N} \sum_{\mu=1}^N W(\mbf{X}_{\mu}).
\label{eqn:Z_AIS}
\end{align}
Using equation (\ref{eqn:Z_AIS}), the free energy is approximated as
\begin{align}
F \approx F_{\mrm{AIS}}(S_T)&:= -\ln Z_{\mrm{AIS}}(S_T) \nn
&\>= - \ln Z_0  - \ln \Big(\frac{1}{N}\sum_{\mu=1}^N W(\mbf{X}_{\mu})\Big).
\label{eqn:F_AIS}
\end{align}
$Z_{\mrm{AIS}}(S_T)$ is an unbiased estimator for the true partition function 
because its expectation converges to $Z$: $\mathbb{E}_{T}[Z_{\mrm{AIS}}(S_T)] =Z$, 
where $\mathbb{E}_{T}[\cdots]$ denotes the expectation of the distribution over $S_T = \{\mbf{X}_{\mu}\}$, i.e., 
\begin{align}
\mathbb{E}_{T}[\cdots]:=\Big(\prod_{\mu=1}^N\sum_{\mbf{X}_{\mu}}T(\mbf{X}_{\mu})\Big)(\cdots).
\label{eqn:expectation_S_T}
\end{align}
However, $F_{\mrm{AIS}}(S_T)$ is not an unbiased estimator for the true free energy~\cite{RAISE2015}. 
Using Jensen's inequality,
\begin{align}
\mathbb{E}_{T}[F_{\mrm{AIS}}(S_T)] &= - \mathbb{E}_{T}[\ln Z_{\mrm{AIS}}(S_T)]\nn
& \geq - \ln \mathbb{E}_{T}[Z_{\mrm{AIS}}(S_T)] = F
\label{eqn:F_AIS_bound}
\end{align}
is obtained, which implies that the expectation of $F_{\mrm{AIS}}(S_T)$ provides an upper bound of the true free energy.

\section{Proposed Method}
\label{sec:proposed_sec}

Suppose that $\bm{x}$ is divided into two mutually disjoint sets: $\bm{v}:= \{v_i \mid i \in V\}$ and $\bm{h}:= \{h_j \mid j \in H\}$, 
i.e., $U = V \cup H$ and $V \cap H = \emptyset$; therefore, $P_k(\bm{x})$ can be considered the joint distribution over $\bm{v}$ and $\bm{h}$: 
$P_k(\bm{x}) = P_k(\bm{v}, \bm{h})$. 
The method discussed in Section \ref{sec:AIS} is regarded as AIS based on this joint distribution. 
In contrast, mAIS proposed in this section is regarded as AIS based on a \textit{marginal} distribution of $P_k(\bm{v}, \bm{h})$.

\subsection{Marginalized annealed importance sampling}
\label{sec:mAIS}

For the sequence of the distributions in equation (\ref{eqn:sequence_distribution}), 
we introduce the sequence of their marginal distributions as  
\begin{align}
\{P_k(\bm{v}) \mid  k = 0,1,\ldots, K\},
\label{eqn:marginal_sequence_distribution}
\end{align}
where $P_k(\bm{v})$ is the marginal distribution of $P_k(\bm{x})$, i.e., 
\begin{align}
P_k(\bm{v}) = \sum_{\bm{h}}P_k(\bm{x})=\frac{1}{Z_k} \exp\big( - E_V(\bm{v}, k)\big),
\label{eqn:marginal_P_k(v)}
\end{align}
where
\begin{align}
E_V(\bm{v}, k):= - \ln \sum_{\bm{h}}\exp\big(- E_k(\bm{x})\big)
\label{eqn:marginal_energy_V}
\end{align}
is the energy function of the marginal distribution. 
From the definition, the partition functions of $P_k(\bm{x})$ and $P_k(\bm{v})$ are the same.

Based on the sequence of the marginal distributions in equation (\ref{eqn:marginal_sequence_distribution}),
in a similar manner to equation (\ref{eqn:annealing_process}), the annealing process from the initial state $\bm{v}^{(1)}$ to the final state $\bm{v}^{(K)}$ can be defined as
\begin{align}
\tau(\bm{V}):= \Big( \prod_{k=1}^{K-1} \tau_k(\bm{v}^{(k+1)} \mid \bm{v}^{(k)})\Big)P_0(\bm{v}^{(1)}),
\label{eqn:annealing_process_mAIS}
\end{align}
where $\bm{V}:= \{\bm{v}^{(k)} \mid k = 1,2,\ldots, K\}$ and $\tau_k(\bm{v}' \mid \bm{v})$ is the transition probability on the marginal distribution in equation (\ref{eqn:marginal_P_k(v)}). 
The transition probability satisfies the following condition: 
\begin{align}
P_k(\bm{v}') = \sum_{\bm{v}} \tau_k(\bm{v}' \mid \bm{v}) P_k(\bm{v}).
\label{eqn:balance_condition_mAIS}
\end{align}
Using almost the same derivation to obtain equation (\ref{eqn:Zk_AIS}), we derive 
\begin{align}
Z_K = Z_0 \sum_{\bm{V}} \Lambda(\bm{V})\tau(\bm{V}),
\label{eqn:Zk_mAIS}
\end{align}
where $\Lambda(\bm{V})$ is the importance weight of mAIS defined as
\begin{align}
\Lambda(\bm{V})&:=\prod_{k=1}^K\lambda_k(\bm{v}^{(k)}),
\label{eqn:mAIS_IW_Lambda}
\end{align}
where
\begin{align}
\lambda_k(\bm{v})&:= \exp\big( - E_V(\bm{v}, k)  + E_V(\bm{v}, k-1)\big).
\label{eqn:mAIS_IW_lambda_k}
\end{align}

In mAIS, the free energy can be evaluated using a technique similar to the derivation of equation (\ref{eqn:F_AIS}).
The sequence of the sample points, $\mbf{V}:=\{\mbf{v}^{(1)}, \mbf{v}^{(2)}, \ldots, \mbf{v}^{(K)}\}$, is generated based on the ancestral sampling on $\tau(\bm{V})$:
\begin{align}
\begin{split}
\mbf{v}^{(1)}&\leftarrow P_0(\bm{v}),\\
\mbf{v}^{(k)}&\leftarrow \tau_{k-1}(\bm{v} \mid \mbf{v}^{(k-1)})\>\> (k = 2,3,\ldots,K).
\end{split}
\label{eqn:mAIS_sampling}
\end{align}
By conducting this ancestral sampling process $N$ times, 
$N$ independent sample sequences $S_{\tau} :=\{\mbf{V}_{\mu} \mid \mu = 1,2,\ldots,N\}$ are obtained, 
and subsequently, using the $N$ sample sequences, equation (\ref{eqn:Zk_mAIS}) is approximated as
\begin{align}
Z \approx Z_{\mrm{mAIS}}(S_{\tau}):=\frac{Z_0}{N} \sum_{\mu=1}^N \Lambda(\mbf{V}_{\mu}).
\label{eqn:Z_mAIS}
\end{align}
Therefore, the free energy is evaluated as
\begin{align}
F \approx F_{\mrm{mAIS}}(S_{\tau})&:= -\ln Z_{\mrm{mAIS}}(S_{\tau}) \nn
&\>= - \ln Z_0  - \ln \Big(\frac{1}{N}\sum_{\mu=1}^N \Lambda(\mbf{V}_{\mu})\Big).
\label{eqn:F_mAIS}
\end{align}
Similar to $Z_{\mrm{AIS}}(S_T)$, $Z_{\mrm{mAIS}}(S_{\tau})$ acts as an unbiased estimator for the true partition function 
because $\mathbb{E}_{\tau}[Z_{\mrm{mAIS}}(S_{\tau})] =Z$,
where $\mathbb{E}_{\tau}[\cdots]$ denotes the expectation of the distribution over $S_{\tau} = \{\mbf{V}_{\mu}\}$, i.e., 
\begin{align}
\mathbb{E}_{\tau}[\cdots]:=\Big(\prod_{\mu=1}^N\sum_{\mbf{V}_{\mu}}\tau(\mbf{V}_{\mu})\Big)(\cdots).
\label{eqn:expectation_S_tau}
\end{align}
Similar to equation (\ref{eqn:F_AIS_bound}), based on Jensen's inequality,
\begin{align}
\mathbb{E}_{\tau}[F_{\mrm{mAIS}}(S_{\tau})]\geq F
\label{eqn:F_mAIS_bound}
\end{align}
can be obtained; therefore, $F_{\mrm{mAIS}}(S_{\tau})$ is not an unbiased estimator for the true free energy as is the case with $F_{\mrm{AIS}}(S_T)$. 

In the aforementioned discussions, 
we have considered mAIS based on the sequence of $P_k(\bm{v})$ (we refer to this mAIS as ``mAIS-V''). 
An opposite mAIS (referred to as ``mAIS-H''), which is based on the sequence of 
\begin{align}
P_k(\bm{h}) &= \sum_{\bm{v}}P_k(\bm{x})= \frac{1}{Z_k} \exp\big( - E_H(\bm{h}, k)\big),
\label{eqn:marginal_P_k(h)}
\end{align}
where
\begin{align}
E_H(\bm{h}, k)&:= - \ln \sum_{\bm{v}}\exp\big(- E_k(\bm{x})\big),
\label{eqn:marginal_energy_H}
\end{align}
can be constructed in the same manner. 
However, we do not need to consider mAIS-H separately 
because the difference between the methods only lies in the way the variables are labelled, 
i.e., mAIS-V is identified with mAIS-H by exchanging the labels of variable sets.  
Therefore, we consider mAIS-V as mAIS in this paper.

Theoretically, mAIS can be applied to any model. 
However, whether it is practical or not strongly depend on the structures of $\{P_k(\bm{x})\}$ 
because mAIS requires the marginal operation, as expressed in equation (\ref{eqn:marginal_P_k(v)}). 
mAIS can be efficiently applied to a bipartite-type MRF (i.e., an MRF defined on a bipartite undirected graph), 
which will be discussed in Section \ref{sec:application_to_BT_MRF}.

Suppose that we have a model that mAIS can be efficiently applied to. 
Therefore, the free energy of the model can be evaluated based on the two methods, i.e., $F_{\mrm{AIS}}(S_T)$ and $F_{\mrm{mAIS}}(S_{\tau})$. 
Intuitively, mAIS seems to be better because the entire sample space of the variables is reduced through the marginalization, 
which is similar to the concept of \textit{Rao--Blackwellization}~\cite{liu2001}.

Here, we briefly explain Rao--Blackwellization.
Suppose that we wish to evaluate the expectation of $f(\bm{x}_{t})$ over a distribution $Q(\bm{x})$:
$m_t := \sum_{\bm{x}} f(\bm{x}_t) Q(\bm{x})$, where $\bm{x}_t \subseteq \bm{x}$. 
Based on a simple sampling approximation (or Monte Carlo integration), it is approximated by
\begin{align}
m_t \approx \frac{1}{M}\sum_{\nu=1}^M f(\mbf{x}_t^{(\nu)}),
\label{eqn:m_t-MCI}
\end{align}
where $\{\mbf{x}^{(\nu)} \mid \nu = 1,2,\ldots, M\}$ is the sample set generated from $Q(\bm{x})$, 
and $\mbf{x}_t^{(\nu)}$ is the corresponding subset of $\mbf{x}^{(\nu)}$. 
Using the decomposition $Q(\bm{x}) = Q(\bm{x}_a \mid \bm{x}_b) Q(\bm{x}_b)$, $m_t$ can be transformed as
\begin{align}
m_t = \sum_{\bm{x}_b} m_t^{\dagger}(\bm{x}_b) Q(\bm{x}_b),
\label{eqn:m_t-conditional}
\end{align}
where $m_t^{\dagger}(\bm{x}_b)$ is the conditional expectation expressed as
\begin{align*}
m_t^{\dagger}(\bm{x}_b):= \sum_{\bm{x}_a}f(\bm{x}_t) Q(\bm{x}_a \mid \bm{x}_b).
\end{align*}
Therefore, based on equation (\ref{eqn:m_t-conditional}), $m_t$ is approximated by 
\begin{align}
m_t \approx \frac{1}{M}\sum_{\nu=1}^M m_t^{\dagger}(\mbf{x}_b^{(\nu)}),
\label{eqn:m_t-RaoBlackwell}
\end{align}
where $\mbf{x}_b^{(\nu)}$ is the corresponding subset of $\mbf{x}^{(\nu)}$.
The Rao--Blackwell theorem guarantees that the estimator in equation (\ref{eqn:m_t-RaoBlackwell}) is more effective than the estimator in equation (\ref{eqn:m_t-MCI}) 
from the perspective of the variance. 
The transformation of equation (\ref{eqn:m_t-conditional}) is called Rao--Blackwellization. 
The applications of Rao--Blackwellization are widely investigated; 
for example, its application to MRFs, spatial Monte Carlo integration, has been developed~\cite{Yasuda2015,Yasuda2021,sekimoto2021}. 

The relation between equations (\ref{eqn:m_t-MCI}) and (\ref{eqn:m_t-RaoBlackwell}) looks similar to that between AIS and mAIS. 
Thus, we can expect mAIS to be more effective. 
In fact this expectation can be justified under a certain condition, which is discussed in the next section.


\subsection{Statistical efficiency of mAIS}
\label{sec:AIS_vs_mAIS}

First, we compare the statistical efficiencies of the two estimators for the partition function, i.e., $Z_{\mrm{AIS}}(S_T)$ and $Z_{\mrm{mAIS}}(S_{\tau})$.
The variance of  $Z_{\mrm{AIS}}(S_T)$ is 
\begin{align}
\mathbb{V}_{T}[Z_{\mrm{AIS}}(S_T)] = \frac{1}{N}\Big(Z_0^2 \sum_{\bm{X}}W(\bm{X})^2T(\bm{X}) - Z^2\Big),
\label{eqn:variance_Z_AIS}
\end{align}
where $\mathbb{V}_{T}[A]:=\mathbb{E}_{T}[A^2] - \mathbb{E}_{T}[A]^2$. 
The variance of  $Z_{\mrm{mAIS}}(S_{\tau})$ is
\begin{align}
\mathbb{V}_{\tau}[Z_{\mrm{mAIS}}(S_{\tau})] = \frac{1}{N}\Big( Z_0^2\sum_{\bm{V}}\Lambda(\bm{V})^2\tau(\bm{V}) - Z^2\Big),
\label{eqn:variance_Z_mAIS}
\end{align}
where $\mathbb{V}_{\tau}[A]:=\mathbb{E}_{\tau}[A^2] - \mathbb{E}_{\tau}[A]^2$. 
As discussed  in Sections \ref{sec:AIS} and \ref{sec:mAIS}, both estimators are unbiased; 
therefore, the estimator with smaller variance is more effective. 

\begin{figure}[tb]
\centering
\includegraphics[height=3cm]{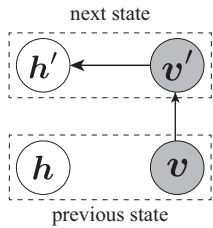}
\caption{State transition, from the previous to next states, according to equation (\ref{eqn:transition_assumption}).}
\label{fig:transition_scheme}
\end{figure}

Assume that the transition probability of AIS is expressed as
\begin{align}
T_k(\bm{x}' \mid \bm{x}) = P_k(\bm{h}' \mid \bm{v}') \tau_k(\bm{v}' \mid \bm{v}),
\label{eqn:transition_assumption}
\end{align}
where $\tau_k(\bm{v}' \mid \bm{v})$ is the transition probability of mAIS satisfying equation (\ref{eqn:balance_condition_mAIS}), and
\begin{align}
P_k(\bm{h}\mid \bm{v}) = \frac{P_k(\bm{x})}{\sum_{\bm{h}} P_k(\bm{x})}
=\frac{\exp\big(-E_k(\bm{x})\big)}{\exp \big( - E_V(\bm{v}, k) \big)}
\label{eqn:conditional_dist_H|V}
\end{align}
is the conditional distribution on $P_k(\bm{x})$. 
According to equation (\ref{eqn:transition_assumption}), the state transition from $\bm{x}$ to $\bm{x}'$ is performed as follows: 
first, the state of $\bm{v}$ is updated to $\bm{v}'$ according to  $\tau_k(\bm{v}' \mid \bm{v})$; 
subsequently, the state of $\bm{h}$ is updated to $\bm{h}'$ according to $P_k(\bm{h}' \mid \bm{v}')$ (see figure \ref{fig:transition_scheme}).
The use of the transition probability of equation (\ref{eqn:transition_assumption}) is accepted in AIS because it satisfies the condition of equation (\ref{eqn:balance_condition}) as follows:
\begin{align*}
&\sum_{\bm{x}}P_k(\bm{h}' \mid \bm{v}') \tau_k(\bm{v}' \mid \bm{v}) P_k(\bm{x})\nn
&=P_k(\bm{h}' \mid \bm{v}') \underbrace{\sum_{\bm{v}}\tau_k(\bm{v}' \mid \bm{v}) P_k(\bm{v})}_{P_k(\bm{v}')} = P_k(\bm{x}').
\end{align*} 
Based on this assumption, the following theorem can be obtained.
\begin{theorem}
$Z_{\mrm{AIS}}(S_T)$ and $Z_{\mrm{mAIS}}(S_{\tau})$ are unbiased estimators for $Z$ defined in equations (\ref{eqn:Zk_AIS}) and (\ref{eqn:Zk_mAIS}), respectively.  
For the two estimators, the inequality
\begin{align*}
\mathbb{V}_{T}[Z_{\mrm{AIS}}(S_T)] \geq \mathbb{V}_{\tau}[Z_{\mrm{mAIS}}(S_{\tau})] 
\end{align*}
always holds, if the transition probabilities of AIS and mAIS satisfy the condition expressed in equation (\ref{eqn:transition_assumption}).
\label{theo:variance_comparison}
\end{theorem}
Based on this theorem, it is ensured that $Z_{\mrm{mAIS}}(S_{\tau})$ is statistically more efficient. 
The proof of this theorem is described in Appendix \ref{app:proof-1}. 
As mentioned in the final part of Appendix \ref{app:proof-1}, the claim of this theorem is essentially the same as that of the Rao--Blackwell theorem.

Next, the statistical efficiencies of the two estimators for the free energy, i.e., $F_{\mrm{AIS}}(S_T)$ and $F_{\mrm{mAIS}}(S_{\tau})$, are compared. 
As expressed in equations (\ref{eqn:F_AIS_bound}) and (\ref{eqn:F_mAIS_bound}), 
$\mathbb{E}_{T}[F_{\mrm{AIS}}(S_{T})]$ and $\mathbb{E}_{\tau}[F_{\mrm{mAIS}}(S_{\tau})]$ are upper bounds of the true free energy. 
Based on the assumption of equation (\ref{eqn:transition_assumption}), the following theorem can be obtained. 
\begin{theorem}
$F_{\mrm{AIS}}(S_T)$ and $F_{\mrm{mAIS}}(S_{\tau})$ are biased estimators for $F$ 
defined in equations (\ref{eqn:F_AIS}) and (\ref{eqn:F_mAIS}), respectively.  
For the two estimators, the inequality
\begin{align*}
\mathbb{E}_{T}[F_{\mrm{AIS}}(S_{T})] \geq \mathbb{E}_{\tau}[F_{\mrm{mAIS}}(S_{\tau})] \geq F
\end{align*}
always holds, if the transition probabilities of AIS and mAIS satisfy the condition expressed in equation (\ref{eqn:transition_assumption}).
\label{theo:F_bound}
\end{theorem}
Based on this theorem, although $F_{\mrm{AIS}}(S_T)$ and $F_{\mrm{mAIS}}(S_{\tau})$ are biased, 
the bias of $F_{\mrm{mAIS}}(S_{\tau})$ is ensured to be smaller. 
The proof of this theorem is presented in Appendix \ref{app:proof-2}.
The aforementioned two theorems do not depend on the details of the design of $E_k(\bm{x})$ and the values of $N$ and $K$; 
only the condition in equation (\ref{eqn:transition_assumption}) is required.

We have proved that mAIS is statistically more efficient than AIS with regard to the partition function and free energy 
when the condition of equation (\ref{eqn:transition_assumption}) is satisfied. 
The remaining issue is determining whether this condition is satisfied in practical situations. 
This condition significantly limits the modeling of $T_k(\bm{x}' \mid \bm{x})$; 
e.g., the standard (synchronous or asynchronous) Gibbs sampling on $P_k(\bm{x})$ does not satisfy the condition 
because it uses the states of all variables to obtain the subsequent states. 
As explained in Section \ref{sec:application_to_BT_MRF}, the state transition according to equation (\ref{eqn:transition_assumption}) can be natural and practical 
when $P_k(\bm{x})$ is a bipartite-type MRF.

In Theorems \ref{theo:variance_comparison} and \ref{theo:F_bound}, 
the condition of equation (\ref{eqn:transition_assumption}) is not a necessary condition but a sufficient condition. 
This implies the possibility of the relaxation of the condition. 
However, it is an open problem.

\subsection{Application to bipartite-type MRFs}
\label{sec:application_to_BT_MRF}

\begin{figure}[tb]
\centering
\includegraphics[height=1.7cm]{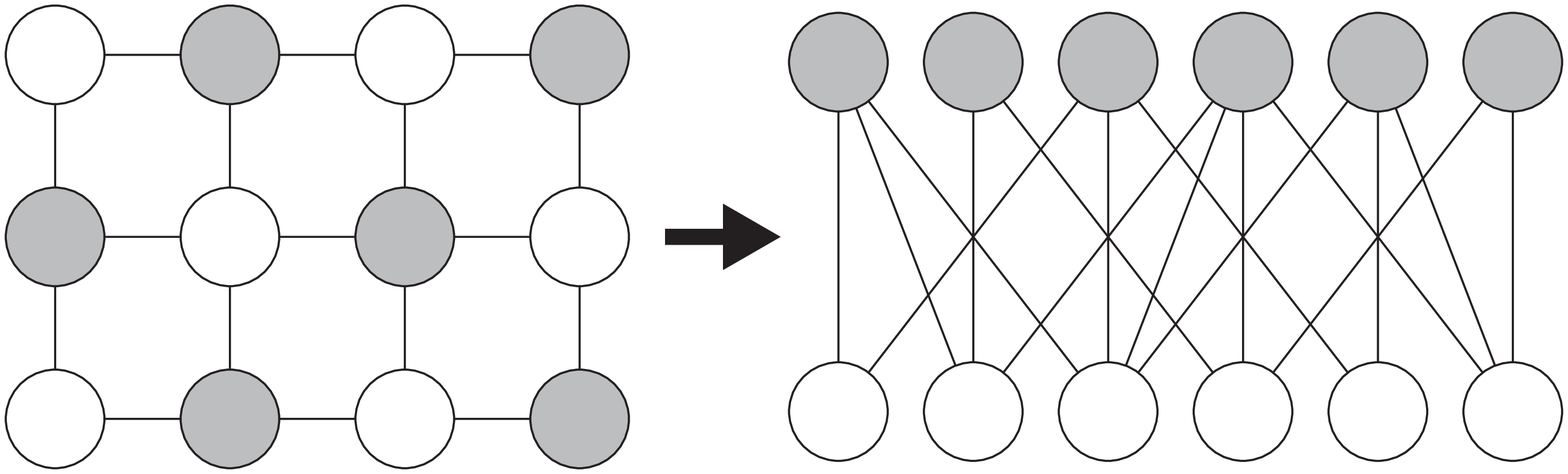}
\caption{Square-grid graph can be regarded as a bipartite graph.}
\label{fig:grid_to_bipartite}
\end{figure}

The proposed mAIS is applicable when the marginal operation in equation (\ref{eqn:marginal_P_k(v)}) is feasible.
Owing to this restriction, the range of application of mAIS is limited. 
As discussed below, mAIS is practical on bipartite-type MRFs,  
which include important applications such as RBM and DBM (a DBM  can be observed as a bipartite-type MRF~\cite{Gao2017}). 
Moreover, a square-grid-type MRF is regarded as a bipartite-type MRF (see Fig. \ref{fig:grid_to_bipartite}).

Suppose that $P_k(\bm{x})$ is a bipartite-type MRF and that $\bm{v}$ and $\bm{h}$ are the variables of each layers. 
In this case, the energy function $E_k(\bm{x})$ can be expressed as
\begin{align}
E_k(\bm{x}) &= -\sum_{i \in V}\phi_i^{(k)}(v_i) - \sum_{j \in H}\phi_j^{(k)}(h_j) \nn
\aleq - \sum_{i \in V}\sum_{j \in H}\psi_{i,j}^{(k)}(v_i,h_j),
\label{eqn:bipartite_energy_k}
\end{align}
where $\phi_i^{(k)}$ and $\psi_{i,j}^{(k)}$ are one- and two-variable functions determined based on the design of $\{P_k(\bm{x})\}$;  
the marginal operation in equation (\ref{eqn:marginal_P_k(v)}) is feasible and leads to 
\begin{align}
E_V(\bm{v}, k)&=-\sum_{i \in V}\phi_i^{(k)}(v_i)  - \sum_{j \in H}\ln \sum_{h_j}\exp \Big(\phi_j^{(k)}(h_j) \nn
\aleq+ \sum_{i \in V}\psi_{i,j}^{(k)}(v_i,h_j)\Big).
\label{eqn:bipartite_marginal_energy_V}
\end{align}

\begin{figure}[tb]
\centering
\includegraphics[height=2cm]{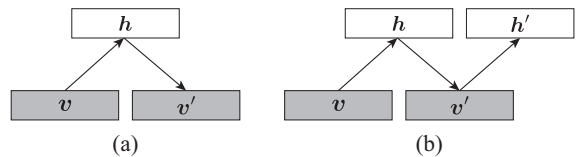}
\caption{State transitions on a bipartite-type MRF based on the one-sample approximation of equations (\ref{eqn:tau_k_bipartite}) and (\ref{eqn:T_k_bipartite}): 
the transitions of (a) mAIS and (b) AIS. 
The transitions between the layers (illustrated by the arrows) are the blocked Gibbs sampling based on $P_k(\bm{h}\mid \bm{v})$ and $P_k(\bm{v}\mid \bm{h})$.}
\label{fig:blocked_Gibbs}
\end{figure}

On the bipartite-type MRF, the layer-wise blocked Gibbs sampling based on the conditional distributions $P_k(\bm{h}\mid \bm{v})$ and $P_k(\bm{v}\mid \bm{h})$ can be easily implemented, 
where $P_k(\bm{h}\mid \bm{v})$ is the conditional distribution presented in equation (\ref{eqn:conditional_dist_H|V}) and
\begin{align}
P_k(\bm{v}\mid \bm{h}) = \frac{P_k(\bm{x})}{\sum_{\bm{v}} P_k(\bm{x})}
=\frac{\exp\big(-E_k(\bm{x})\big)}{\exp \big( - E_H(\bm{h}, k) \big)},
\label{eqn:conditional_dist_V|H}
\end{align}
because $v_i$s are conditionally independent from each other in $P_k(\bm{v}\mid \bm{h})$
and $h_j$s are also conditionally independent from each other in $P_k(\bm{h}\mid \bm{v})$, i.e.,  
\begin{align*}
P_k(\bm{v}\mid \bm{h}) &\propto \prod_{i \in V} \exp\Big(\phi_i^{(k)}(v_i) + \sum_{j \in H}\psi_{i,j}^{(k)}(v_i,h_j)\Big),\nn
P_k(\bm{h}\mid \bm{v}) &\propto \prod_{j \in H} \exp\Big(\phi_j^{(k)}(h_j) + \sum_{i \in V}\psi_{i,j}^{(k)}(v_i,h_j)\Big).
\end{align*}
Based on the layer-wise blocked Gibbs sampling, the transition probability of mAIS can be modeled as
\begin{align}
\tau_k(\bm{v}' \mid \bm{v})=\sum_{\bm{h}}P_k(\bm{v}'\mid \bm{h})P_k(\bm{h}\mid \bm{v}),
\label{eqn:tau_k_bipartite}
\end{align}
which satisfies the condition in equation (\ref{eqn:balance_condition_mAIS}) as follows: 
\begin{align*}
\sum_{\bm{v}} \sum_{\bm{h}}P_k(\bm{v}'\mid \bm{h})P_k(\bm{h}\mid \bm{v})P_k(\bm{v})= P_k(\bm{v}').
\end{align*}
This transition probability can be considered as a collapsed Gibbs sampling~\cite{CGibbs1994}. 
From equations (\ref{eqn:transition_assumption}) and (\ref{eqn:tau_k_bipartite}), 
when the transition probability of AIS is modeled by
\begin{align}
T_k(\bm{x}' \mid \bm{x}) = P_k(\bm{h}'\mid \bm{v}')\sum_{\bm{h}}P_k(\bm{v}'\mid \bm{h})P_k(\bm{h}\mid \bm{v}),
\label{eqn:T_k_bipartite}
\end{align}
the theoretical results mentioned in Section \ref{sec:AIS_vs_mAIS} (Theorems \ref{theo:variance_comparison} and \ref{theo:F_bound}) are applicable.
Approximating the expectations in equations  (\ref{eqn:tau_k_bipartite}) and (\ref{eqn:T_k_bipartite}) using the sampling approximation with one sample point
leads to the widely used sampling procedure, i.e., the blocked Gibbs sampling, on bipartite-type MRFs (see figure \ref{fig:blocked_Gibbs}).

\section{Numerical Experiments}
\label{sec:experiment}

The performance of mAIS (i.e., mAIS-V) is examined using numerical experiments on an RBM whose energy function is defined as 
\begin{align}
E(\bm{x})=-\frac{1}{T}\Big(\sum_{i \in V} b_i v_i + \sum_{j \in H}c_j h_j + \sum_{i \in V}\sum_{j \in H}w_{i,j}v_ih_j\Big),
\label{eqn:RBM}
\end{align}
where $v_i , h_j \in \{-1,+1\}$ and $T > 0$ is the temperature of the RBM. 
The temperature controls the complexity of the distribution: a lower $T$ expresses a higher-clustered multimodal distribution.  
On the RBM, the distribution sequence, $\{P_k(\bm{x})\}$, is designed according to equation (\ref{eqn:P_k(x)_beta_k}), 
where $\beta_{k+1} = \beta_k + 1/K$ and the initial distribution $P_0(\bm{x})$ is fixed to a uniform distribution; 
therefore, in this case, 
$E_k(\bm{x})$ and $E_V(\bm{v}, k)$ in equations (\ref{eqn:bipartite_energy_k}) and (\ref{eqn:bipartite_marginal_energy_V})  are
\begin{align*}
E_k(\bm{x}) = -\frac{\beta_k}{T}\Big(\sum_{i \in V} b_i v_i + \sum_{j \in H}c_j h_j + \sum_{i \in V}\sum_{j \in H}w_{i,j}v_ih_j \Big)
\end{align*}
and 
\begin{align*}
E_V(\bm{v}, k) &= -\frac{\beta_k}{T}\sum_{i \in V} b_i v_i \nn
\aleq- \sum_{j \in H} \ln 2\cosh \frac{\beta_k}{T}\Big(c_j +\sum_{i \in V} w_{i,j} v_i\Big),
\end{align*}
respectively.   

\subsection{Quantitative investigation of theoretical results}
\label{sec:exper-1}

The theoretical results obtained in section \ref{sec:AIS_vs_mAIS} revealed the qualitatively effectiveness of mAIS. 
We investigate the quantitatively effectiveness of mAIS using numerical experiments.
In the experiments, $b_i$s and $c_j$s were independently drawn from a uniform distribution in $[-0.001, +0.001]$, 
and $w_{i,j}$s are independently drawn from a normal distribution whose mean is zero and variance is $1 / n$; 
the sizes of the sample set and $\bm{v}$ were fixed to $N = 1000$ and $|V| = 20$.  
The blocked Gibbs sampling shown in figure \ref{fig:blocked_Gibbs} was used for the state transitions of AIS and mAIS.

\begin{figure}[t]
\centering
\includegraphics[height=4cm]{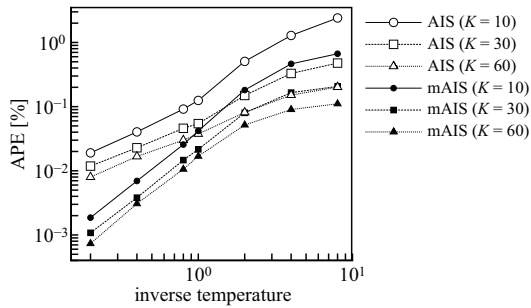}
\caption{APEs versus $1/T$ for $K = 10$, 30 and 60. 
These plots represent the average values over 1000 experiments.}
\label{fig:ape_vs_T}
\end{figure}
Figure \ref{fig:ape_vs_T} shows the absolute percentage error (APE) between the true free energy, $f := F/n$, 
and its approximation, $f_{\mrm{app}} := F_{\mrm{app}} / n$, obtained using AIS or mAIS, against the inverse temperature $1 / T$, 
where the APE is defined as
\begin{align}
\mrm{APE}:= 100 \times \frac{| f - f_{\mrm{app}}|}{|f|} \> [\%].
\label{eqn:ape}
\end{align}
In this experiment, the size of $\bm{h}$ is fixed to $|H| = 40$. 
We observe that the APEs obtained using mAIS are always lower than those obtained using AIS, as supported by Theorem \ref{theo:variance_comparison}.
mAIS is particularly effective in the high-temperature region.

\begin{table*}[t]
\caption{Comparison of the true free energy $f$ with the trial averages of its approximations, $\mathbb{E}_{\mrm{trial}}[f_{\mrm{app}}]$, 
obtained using AIS and mAIS, for several $T$ and $K$ values. 
The values are the average values obtained over 1000 experiments.}
\begin{tabular}{cccccccc}
\multicolumn{1}{l}{} & \multicolumn{1}{l}{} & \multicolumn{3}{c}{AIS} & \multicolumn{3}{c}{mAIS} \\ \cline{3-8} 
 &  & \multicolumn{3}{c|}{$K$} & \multicolumn{3}{c}{$K$} \\ \cline{3-8} 
$1/T$ & true & 10 & 30 & \multicolumn{1}{c|}{60} & 10 & 30 & 60 \\ \hline
0.2 & $-0.69759$ & $-0.69759$ & $-0.69759$ & \multicolumn{1}{c|}{$-0.69759$} & $-0.69759$ & $-0.69759$ &$ -0.69759$ \\
0.4 & $-0.71084$ & $-0.71083$ & $-0.71084$ & \multicolumn{1}{c|}{$-0.71084$} & $-0.71084$ & $-0.71084$ & $-0.71084$  \\
0.8 & $-0.76376$ & $-0.76373$ & $-0.76375$ & \multicolumn{1}{c|}{$-0.76375$} & $-0.76375$ & $-0.76376$ & $-0.76376$  \\
1    & $-0.80306$ & $-0.80300$ & $-0.80305$ & \multicolumn{1}{c|}{$-0.80305$} & $-0.80305$ & $-0.80306$ & $-0.80306$  \\
2    & $-1.10992$ & $-1.10782$ & $-1.10977$ & \multicolumn{1}{c|}{$-1.10987$} & $-1.10963$ & $-1.10987$ & $-1.10990$  \\
4    & $-1.95593$ & $-1.93328$ & $-1.95345$ & \multicolumn{1}{c|}{$-1.95545$} & $-1.95143$ & $-1.95535$ & $-1.95575$  \\
8    & $-3.80281$ & $-3.70846$ & $-3.78813$ & \multicolumn{1}{c|}{$-3.79920$} & $-3.78087$ & $-3.79925$ & $-3.80186$  \\ \hline
\end{tabular}
\label{tab:free_energy}
\end{table*}
Next, we compare the true free energy $f$ with the trial averages of its approximations, $\mathbb{E}_{\mrm{trial}}[f_{\mrm{app}}]$, 
where $\mathbb{E}_{\mrm{trial}}[\cdots]$ was estimated based on the average of 30 trials. 
The trial average can be regarded as the approximation of $\mathbb{E}_{T}[\cdots]$ in AIS or $\mathbb{E}_{\tau}[\cdots]$ in mAIS.   
The results are listed in table \ref{tab:free_energy}. 
In this experiment, the size of $\bm{h}$ was fixed to $|H| = 40$. 
The estimates of AIS and mAIS are higher than the true free energy 
and the estimates of mAIS are lower than those of AIS, as supported by Theorem \ref{theo:F_bound}.

\subsection{Dependency on the size ratio of two layers}
\label{sec:exper-2}

\begin{figure*}[t]
\centering
\includegraphics[height=4cm]{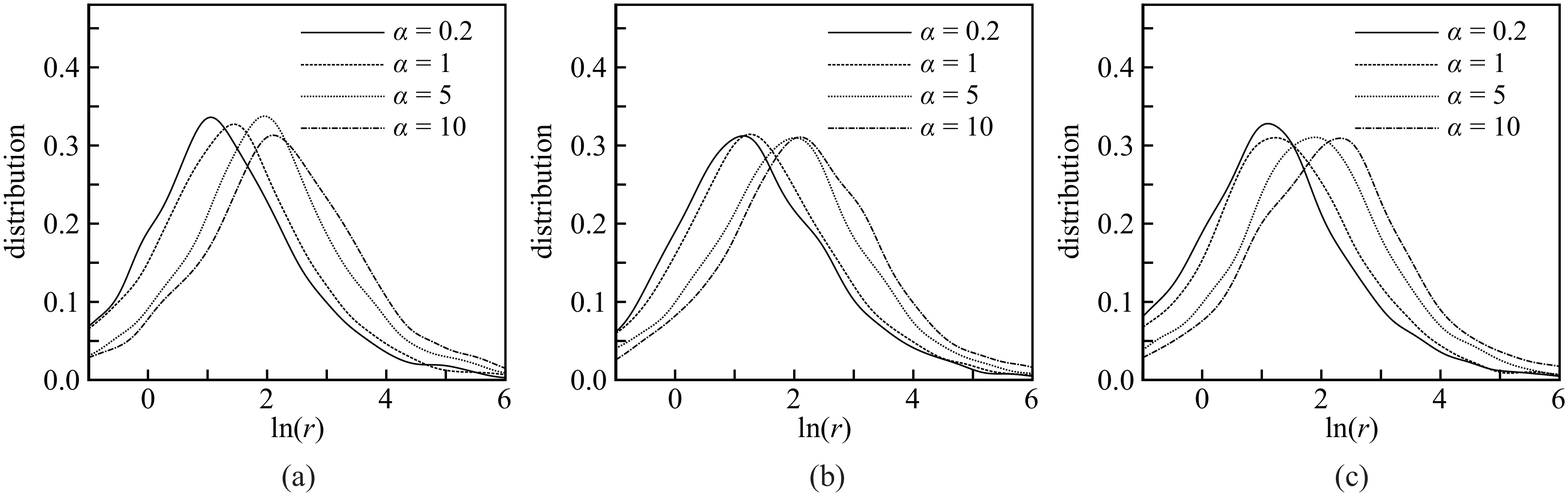}
\caption{Distributions of $\ln r$ for several $\alpha$ values when $1/ T = 0.5$: (a) $K = 10$, (b) $K = 30$, and (c) $K = 60$.}
\label{fig:dist_beta0.5}
\end{figure*}

\begin{figure*}[t]
\centering
\includegraphics[height=4cm]{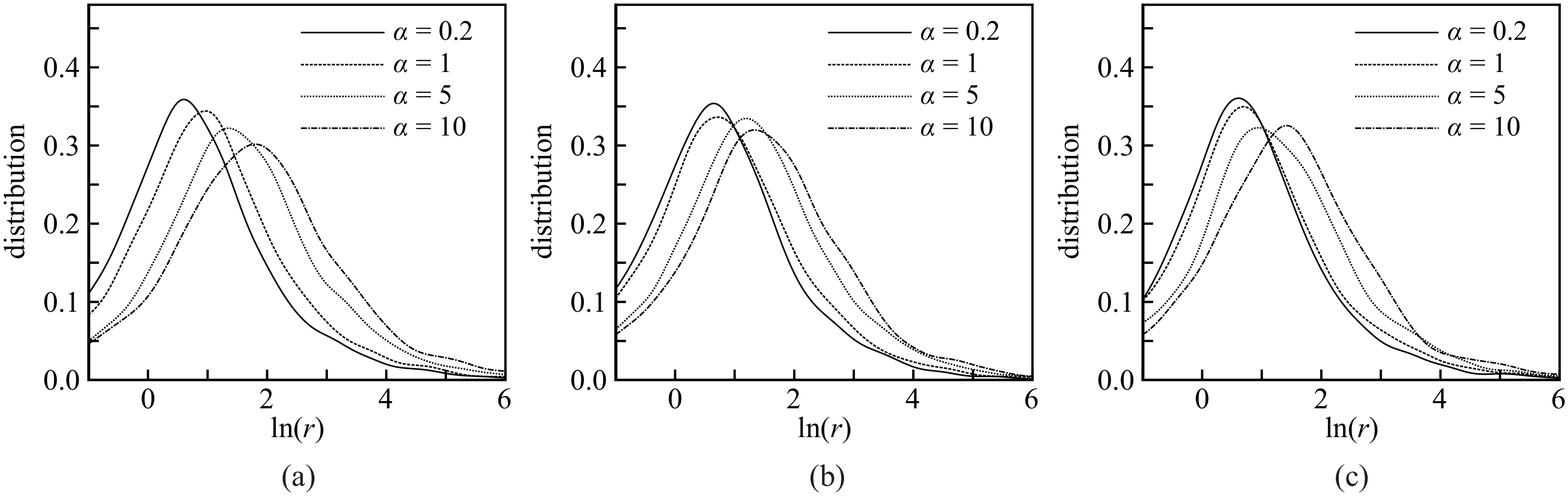}
\caption{Distributions of $\ln r$ for several $\alpha$ values when $1/ T = 1$: (a) $K = 10$, (b) $K = 30$, and (c) $K = 60$.}
\label{fig:dist_beta1}
\end{figure*}

\begin{figure*}[t]
\centering
\includegraphics[height=4cm]{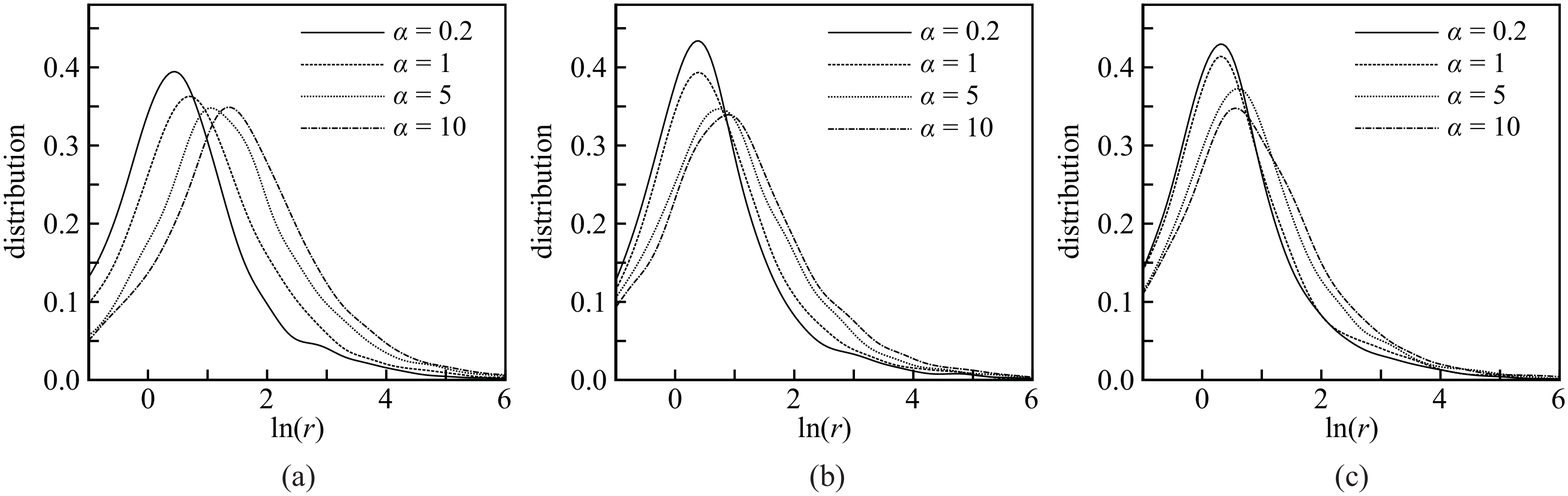}
\caption{Distributions of $\ln r$ for several $\alpha$ values when $1/ T = 2$: (a) $K = 10$, (b) $K = 30$, and (c) $K = 60$.}
\label{fig:dist_beta2}
\end{figure*}

We investigate the relative accuracy of mAIS compared with AIS for fraction $\alpha := |H| / |V|$. 
The experiments in section \ref{sec:exper-1} correspond to the cases of $\alpha = 2$. 
Intuitively, mAIS is more efficient for a larger $\alpha$ 
because, as $\alpha$ increases, the fraction of the size of remaining variables through marginalization to $n$ reduces  
(in other words, the dimension of space evaluated by the sampling approximation relatively shrinks). 
However, the theoretical results obtained in section \ref{sec:AIS_vs_mAIS} do not directly support this assumption. 
Thus, we check the relative accuracy for several $\alpha$ values using numerical experiments on the RBM.
In the experiments, the relative accuracy is measured by the ratio of the APEs obtained using AIS and mAIS, i.e., 
\begin{align*}
r := \frac{\text{APE of AIS} }{ \text{APE of mAIS}} = \frac{|f - f_{\mrm{AIS}}|}{|f - f_{\mrm{mAIS}}|}.
\end{align*}
The relative accuracy increases as $r$ increases. 
In the experiment, the sizes of the sample set and $\bm{v}$ were fixed to $N = 1000$ and $|V| = 20$; 
the parameter setting of the RBM was the same as that in section \ref{sec:exper-1}, and the blocked Gibbs sampling was used for the state transitions of AIS and mAIS.  
The distributions of $\ln r$ for $1/T = 0.5$, $1$, and $2$ are shown in figures \ref{fig:dist_beta0.5}--\ref{fig:dist_beta2}, respectively. 
Each distribution in these figures was created based on the kernel density estimation~\cite{Bishop2006} using the results obtained from 4000 experiments, 
in which the Gaussian kernel with a bandwidth (or smoothing parameter) of 0.25 was used. 
In all cases, the distributions transit to the positive direction with an increase in $\alpha$,  
which implies that the relative accuracy monotonically improves as $\alpha$ increases. 
This result implies that we should use mAIS-V when $\alpha > 1$ and mAIS-H when $\alpha < 1$.

\subsection{Convergence property for $K$}

\begin{figure}[h]
\centering
\includegraphics[height=1.8cm]{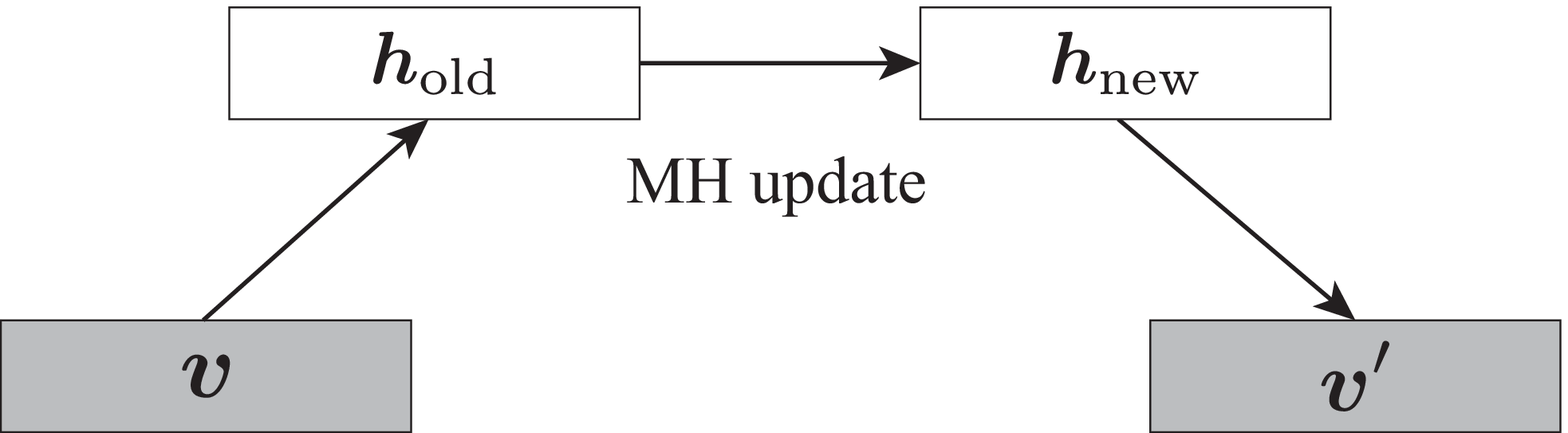}
\caption{Illustration of Roussel's sampling method~\cite{Roussel2021}. The transition from $\bm{h}_{\mrm{old}}$ to $\bm{h}_{\mrm{new}}$ is performed using 
the MH method on $\bm{h}$ space. The transitions between the two layers are the same as the blocked Gibbs sampling. 
This sampling procedure can be used as $\tau_k(\bm{v}' \mid \bm{v})$.}
\label{fig:Roussel}
\end{figure}

The experiments in sections \ref{sec:exper-1} and \ref{sec:exper-2} were conducted for relatively small $K$ values to emphasize the performance difference between AIS and mAIS 
under less-than-ideal conditions. 
However, in practice, $K$ is set to a sufficient large value to obtain precise estimations. 
To obtain precise estimations, the value of $K$ should be set to a value larger than the mixing time (or relaxation time). 
The mixing time tends to increase as the complexity of distribution increases.
In this section, the convergence property of APE in equation (\ref{eqn:ape}) on a wider range of $K$ is demonstrated using numerical experiments.

The detailed discussions for the mixing time in RBM have been provided from both inference and learning perspectives~\cite{Roussel2021,Decelle2021_NIPS}.
Roussel \textit{et al.} proved that the standard blocked Gibbs sampling shown in figure \ref{fig:blocked_Gibbs} is not efficient in RBMs having high clustered distributions from the perspective of the mixing time, 
and they proposed a more efficient sampling method by combining the blocked Gibbs sampling and Metropolis--Hastings (MH) method illustrated in figure \ref{fig:Roussel}~\cite{Roussel2021}. 
Roussel's sampling method can be employed as the transition $\tau_k(\bm{v}' \mid \bm{v})$ in our framework. 
For the experiments, we used two different RBMs: the RBM with Gaussian interactions which is the same model used in sections \ref{sec:exper-1} and \ref{sec:exper-2}, 
and the RBM with Hopfield-type interactions in which $w_{i,j} = \xi_i^{(j)} / \sqrt{|V|}$, where $\xi_i^{(j)} \in \{-1,+1\}$ was determined uniformly at random. 
The Hopfield-type interactions can exhibit a clustered distribution~\cite{Roussel2021}. 
In both RBMs, the bias parameters, $b_i$s and $c_j$s, were independently drawn from a uniform distribution in $[-0.001, +0.001]$, and $|V| = 20$ and $|H| = 40$ were fixed.  
The size of the sample set was fixed to $N = 10000$.  

\begin{figure*}[t]
\centering
\includegraphics[height=4cm]{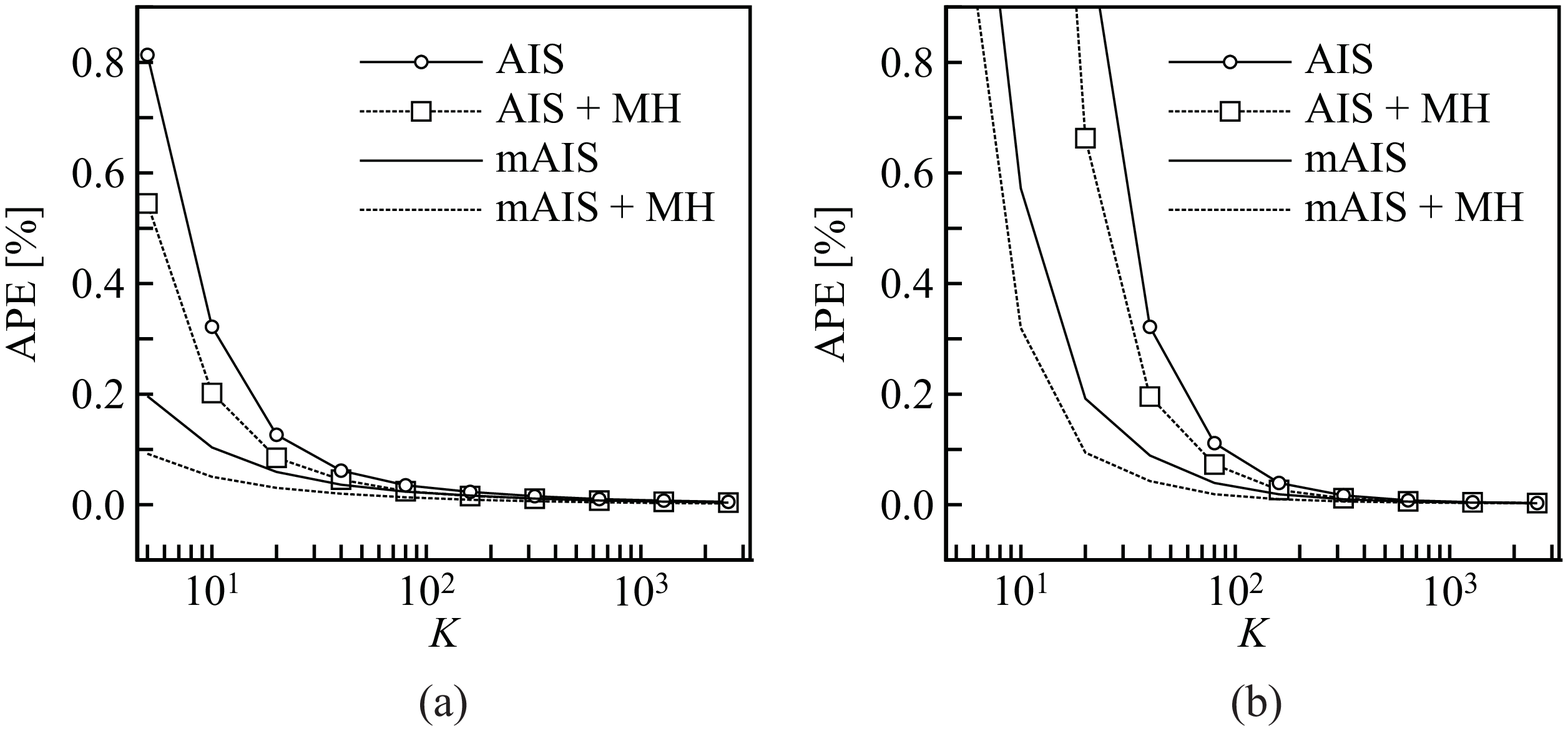}
\caption{APEs versus $K$ on the RBM with the Gaussian interactions: (a) $1/T = 2$ and (b) $1/T = 20$. 
These plots represent the average values over 1000 experiments.}
\label{fig:K_gauss}
\end{figure*}

\begin{figure*}[t]
\centering
\includegraphics[height=4cm]{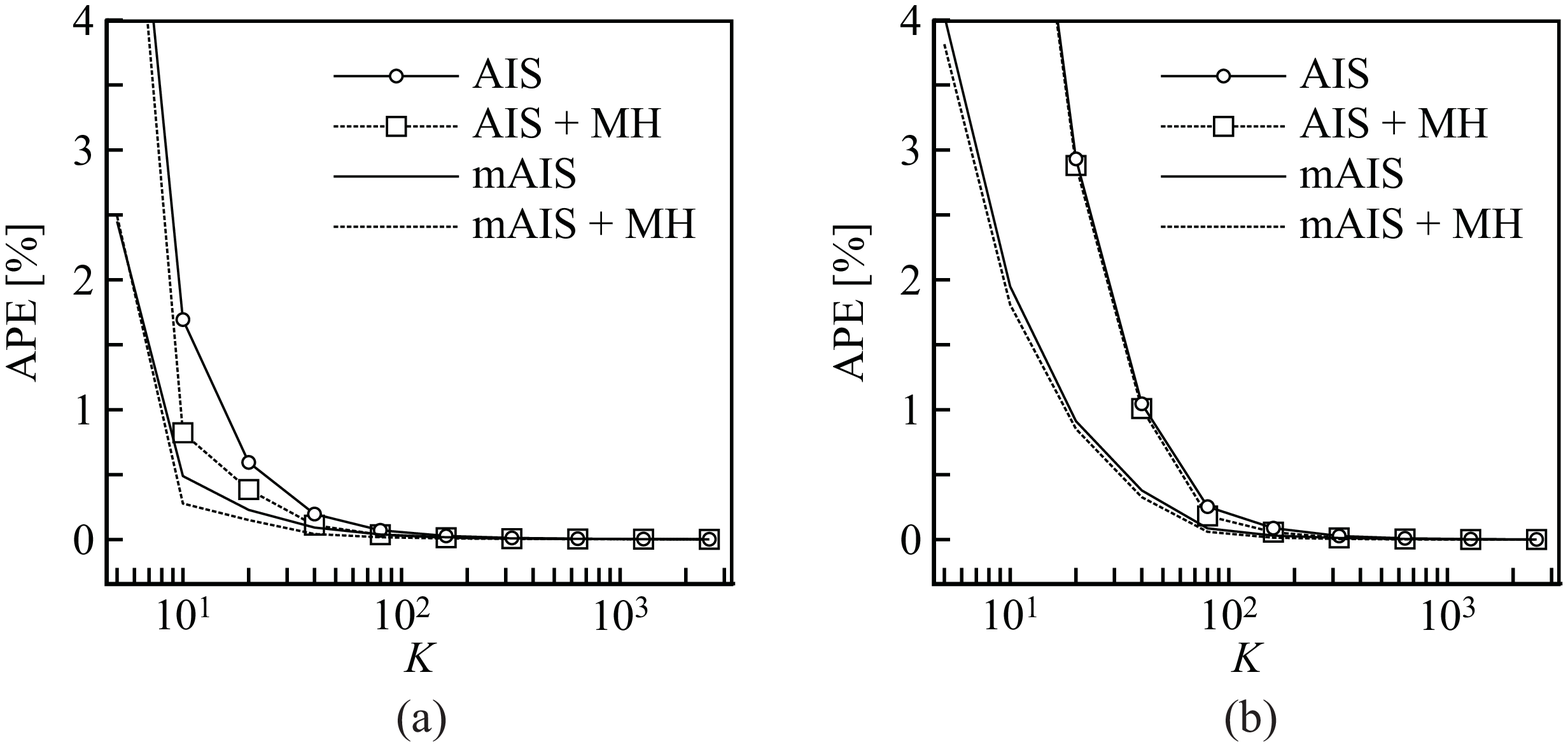}
\caption{APEs versus $K$ on the RBM with the Hopfield-type interactions: (a) $1/T = 2$ and (b) $1/T = 20$. 
These plots represent the average values over 1000 experiments.}
\label{fig:K_hp}
\end{figure*}

Figures \ref{fig:K_gauss} and \ref{fig:K_hp} show the behavior of the APE for the increase in $K$. 
In the figures, ``AIS'' and ``mAIS'' indicate the results obtained using the standard blocked Gibbs sampling, 
while ``AIS+MH'' and ``mAIS+MH'' indicate the results obtained using Roussel's sampling method. 
mAIS exhibits faster convergence than AIS in all experiments, 
and Roussel's sampling method improves the convergence properties of both AIS and mAIS, although the improvement is very small in figure \ref{fig:K_hp}(b). 
mAIS+MH achieved the best performance the best in all experiments. 
The convergence speed of mAIS seems to be higher than that of AIS+MH; 
this implies that the improvement by mAIS is more effective than that by Roussel's sampling method. 
Finally, we comment on the computational efficiency of the four methods: AIS, AIS+MH, mAIS, and mAIS+MH.
The computational cost of the four methods have the same order; they are $O(NK |V| |H|)$. 
However, mAIS is remarkably faster than AIS+MH with regard to the CPU time (around 10 times faster in our implementation). 
The MH procedure is the bottleneck in the Roussel's sampling method. 
Thus, mAIS+MH is as slow as AIS+MH.

\section{Summary and Discussion}
\label{sec:summary}

In this study, we have proposed mAIS, which is identified as AIS on a marginalized model. 
The proposed method is regarded as a special case of AIS, which can be used when a partial marginalization can be exactly evaluated, and it is not an improvement method of AIS. 
This study therefore contributes to several related studies for AIS such as those listed in section \ref{sec:intro}.  
Two important theorems for mAIS (i.e., Theorems \ref{theo:variance_comparison} and \ref{theo:F_bound}) have been obtained: 
when the transition probabilities of AIS and mAIS satisfy the condition of equation (\ref{eqn:transition_assumption}), 
(a) the partition-function estimator of mAIS is more accurate than that of AIS  
and (b) the bias of the free energy estimator of mAIS is lower than that of AIS.
These results theoretically support the qualitative effectiveness of mAIS. 
Furthermore, the numerical results demonstrated in section \ref{sec:experiment} empirically support the quantitative effectiveness of mAIS.

mAIS can be applied to a model in which a partial marginalization can be exactly evaluated. 
However, the effectiveness of the resultant mAIS in comparison with AIS cannot be theoretically guaranteed 
when the transition probabilities of AIS and mAIS do not satisfy the condition of equation (\ref{eqn:transition_assumption}). 
This condition significantly limits the applicability of the theory obtained in this study.
As mentioned in section \ref{sec:AIS_vs_mAIS}, the condition of equation (\ref{eqn:transition_assumption}) 
is sufficient for Theorems \ref{theo:variance_comparison} and \ref{theo:F_bound}, 
and has been not identified as a necessary condition. 
If it is not necessary, then it may be relaxed.  
A relaxation of the condition needs to be investigated in our future studies. 

\appendix

\section{Proofs}
\label{app:proofs}

This appendix demonstrates the proofs of the two theorems presented in Section \ref{sec:AIS_vs_mAIS}.

\subsection{Proof of Theorem \ref{theo:variance_comparison}}
\label{app:proof-1}

As discussed in Sections \ref{sec:AIS} and \ref{sec:mAIS}, $Z_{\mrm{AIS}}(S_T)$ and $Z_{\mrm{mAIS}}(S_{\tau})$ are the unbiased estimators of $Z$ 
because 
\begin{align*}
\mathbb{E}_{T}[Z_{\mrm{AIS}}(S_{T})] =\mathbb{E}_{\tau}[Z_{\mrm{mAIS}}(S_{\tau})] =Z.
\end{align*}

The relationship between the annealing processes in equations (\ref{eqn:annealing_process}) and (\ref{eqn:annealing_process_mAIS}) is considered.
Based on the assumption of equation (\ref{eqn:transition_assumption}), the marginal distribution of $T(\bm{X})$ can be expressed as
\begin{widetext}
\begin{align*}
\sum_{\bm{H}} T(\bm{X}) 
&= \sum_{\bm{H}} \Big(\prod_{k=1}^{K-1} P_k(\bm{h}^{(k+1)} \mid \bm{v}^{(k+1)})\tau_k(\bm{v}^{(k+1)} \mid \bm{v}^{(k)})\Big)
P_0(\bm{v}^{(1)}, \bm{h}^{(1)})\nn
&=\Big(\prod_{k=1}^{K-1} \sum_{\bm{h}^{(k+1)}}P_k(\bm{h}^{(k+1)} \mid \bm{v}^{(k+1)})\tau_k(\bm{v}^{(k+1)} \mid \bm{v}^{(k)})\Big)
\sum_{\bm{h}^{(1)}}P_0(\bm{v}^{(1)}, \bm{h}^{(1)})
=\Big(\prod_{k=1}^{K-1}\tau_k(\bm{v}^{(k+1)} \mid \bm{v}^{(k)})\Big)P_0(\bm{v}^{(1)}).
\end{align*}
\end{widetext}
Therefore, 
\begin{align}
\tau(\bm{V}) = \sum_{\bm{H}}T(\bm{X}) 
\label{eqn:marginal-relation_T&tau}
\end{align}
is obtained. 
Moreover, the assumption of equation (\ref{eqn:transition_assumption}) leads to
\begin{widetext}
\begin{align}
\frac{T(\bm{X}) }{\tau(\bm{V}) }&=\Big(\prod_{k=1}^{K-1} \frac{P_k(\bm{h}^{(k+1)} \mid \bm{v}^{(k+1)})\tau_k(\bm{v}^{(k+1)} \mid \bm{v}^{(k)})}{\tau_k(\bm{v}^{(k+1)} \mid \bm{v}^{(k)})}\Big)
\frac{P_0(\bm{v}^{(1)}, \bm{h}^{(1)})}{P_0(\bm{v}^{(1)})}
=\prod_{k=1}^{K}P_{k-1}(\bm{h}^{(k)} \mid \bm{v}^{(k)}).
\label{eqn:ratio_T/V}
\end{align}
\end{widetext}
From equations (\ref{eqn:marginal-relation_T&tau}) and (\ref{eqn:ratio_T/V}), the annealing process of AIS can be factorizable as 
\begin{align}
T(\bm{X})  = T(\bm{H} \mid \bm{V}) \tau(\bm{V}),
\label{eqn:relation_T&tau}
\end{align}
where $T(\bm{H} \mid \bm{V}):=\prod_{k=1}^{K}P_{k-1}(\bm{h}^{(k)} \mid \bm{v}^{(k)})$ is the conditional distribution over $\bm{H}$.
Using equation (\ref{eqn:relation_T&tau}), the difference in the variances can be expressed as
\begin{align}
&\mathbb{V}_{T}[Z_{\mrm{AIS}}(S_T)] - \mathbb{V}_{\tau}[Z_{\mrm{mAIS}}(S_{\tau})] \nn
&=\frac{Z_0^2}{N}\Big( \sum_{\bm{X}}W(\bm{X})^2T(\bm{X})-  \sum_{\bm{V}}\Lambda(\bm{V})^2 \tau(\bm{V})\Big)\nn
&=\frac{Z_0^2}{N}\sum_{\bm{V}}\Big(\sum_{\bm{H}}W(\bm{X})^2 T(\bm{H} \mid \bm{V}) - \Lambda(\bm{V})^2\Big) \tau(\bm{V}).
\label{eqn:diff_variances}
\end{align}
From equations (\ref{eqn:AIS_IW_wk}) and (\ref{eqn:mAIS_IW_lambda_k}), 
\begin{align*}
&\lambda_k(\bm{v}^{(k)}) = \frac{\sum_{\bm{h}^{(k)}} \exp(-E_k(\bm{x}^{(k)}) )}
{\sum_{\bm{h}^{(k)}} \exp(-E_{k-1}(\bm{x}^{(k)}) )}\nn
&=\sum_{\bm{h}^{(k)}} \frac{\exp(-E_k(\bm{x}^{(k)}) )}{\exp(-E_{k-1}(\bm{x}^{(k)}) )}
\frac{\exp(-E_{k-1}(\bm{x}^{(k)}) )}{\sum_{\bm{h}^{(k)}} \exp(-E_{k-1}(\bm{x}^{(k)}) )}\nn
&= \sum_{\bm{h}^{(k)}} w_k(\bm{x}^{(k)})P_{k-1}(\bm{h}^{(k)} \mid \bm{v}^{(k)})
\end{align*}
is obtained, which leads to
\begin{align}
\Lambda(\bm{V}) = \sum_{\bm{H}}W(\bm{X}) T(\bm{H} \mid \bm{V}).
\label{eqn:relation_W&Lambda}
\end{align}
Substituting equation (\ref{eqn:relation_W&Lambda}) into (\ref{eqn:diff_variances}) yields  
\begin{align*}
&\mathbb{V}_{T}[Z_{\mrm{AIS}}(S_T)] - \mathbb{V}_{\tau}[Z_{\mrm{mAIS}}(S_{\tau})] \nn
&=\frac{Z_0^2}{N}\sum_{\bm{X}}\big( W(\bm{X}) - \Lambda(\bm{V}) \big)^2T(\bm{X})
\geq 0.
\end{align*}
Therefore, $\mathbb{V}_{T}[Z_{\mrm{AIS}}(S_T)] \geq \mathbb{V}_{\tau}[Z_{\mrm{mAIS}}(S_{\tau})]$ is ensured. $\square$

The above result can be understand based on Rao--Blackwellization. 
$Z_{\mrm{AIS}}(S_T)$ is the simple sampling approximation of equation (\ref{eqn:Zk_AIS}).
From equation (\ref{eqn:relation_T&tau}), equation (\ref{eqn:Zk_AIS}) can be rewritten as
\begin{align*}
Z_K = Z_0 \sum_{\bm{V}}\sum_{\bm{H}}W(\bm{X})T(\bm{H} \mid \bm{V}) \tau(\bm{V}).
\end{align*}
From this equation and equation (\ref{eqn:relation_W&Lambda}), equation (\ref{eqn:Zk_mAIS}) can be seen as the expectation of the conditional expectation of $W(\bm{X})$. 
mAIS, therefore, is identified with Rao--Blackwellization of AIS; 
and therefore, $\mathbb{V}_{T}[Z_{\mrm{AIS}}(S_T)] \geq \mathbb{V}_{\tau}[Z_{\mrm{mAIS}}(S_{\tau})]$ is ensured by the Rao--Blackwell theorem.

\subsection{Proof of Theorem \ref{theo:F_bound}}
\label{app:proof-2}

As shown in equations (\ref{eqn:F_AIS_bound}) and (\ref{eqn:F_mAIS_bound}), 
$\mathbb{E}_{T}[F_{\mrm{AIS}}(S_{T})]$ and $\mathbb{E}_{\tau}[F_{\mrm{mAIS}}(S_{\tau})]$ are upper bounds of the true free energy $F$, 
i.e., $\mathbb{E}_{T}[F_{\mrm{AIS}}(S_{T})] \geq F$ and $\mathbb{E}_{\tau}[F_{\mrm{mAIS}}(S_{\tau})] \geq F$. 
Using equation  (\ref{eqn:relation_T&tau}), $\mathbb{E}_{T}[F_{\mrm{AIS}}(S_T)]$ can be rewritten as
\begin{align}
&\mathbb{E}_{T}[F_{\mrm{AIS}}(S_T)] \nn
&=-\ln Z_0 - \Big(\prod_{\mu=1}^N\sum_{\mbf{X}_{\mu}}T(\mbf{X}_{\mu})\Big) \ln \Big(\frac{1}{N}\sum_{\mu=1}^N W(\mbf{X}_{\mu})\Big)\nn
&=-\ln Z_0 \nn
&- \Big(\prod_{\mu=1}^N\sum_{\mbf{V}_{\mu}}\sum_{\mbf{H}_{\mu}}T(\mbf{H}_{\mu} \mid \mbf{V}_{\mu})\tau(\mbf{V}_{\mu})\Big) \ln \Big(\frac{1}{N}\sum_{\mu=1}^N W(\mbf{X}_{\mu})\Big).
\label{eqn:Jensen_for_theorem}
\end{align}
Using Jensen's inequality, the following inequality is obtained: 
\begin{align}
&\Big(\prod_{\mu=1}^N\sum_{\mbf{V}_{\mu}}\sum_{\mbf{H}_{\mu}}T(\mbf{H}_{\mu} \mid \mbf{V}_{\mu})\tau(\mbf{V}_{\mu})\Big) \ln \Big(\frac{1}{N}\sum_{\mu=1}^N W(\mbf{X}_{\mu})\Big)\nn
&\leq 
\Big(\prod_{\mu=1}^N\sum_{\mbf{V}_{\mu}}\tau(\mbf{V}_{\mu})\Big) \ln \Big(\frac{1}{N}\sum_{\mu=1}^N \sum_{\mbf{H}_{\mu}}T(\mbf{H}_{\mu} \mid \mbf{V}_{\mu})W(\mbf{X}_{\mu})\Big)\nn
&=\Big(\prod_{\mu=1}^N\sum_{\mbf{V}_{\mu}}\tau(\mbf{V}_{\mu})\Big) \ln \Big(\frac{1}{N}\sum_{\mu=1}^N \Lambda(\mbf{V}_{\mu})\Big).
\label{eqn:inequality-1}
\end{align}	
Equation (\ref{eqn:relation_W&Lambda}) is used in the derivation of the last line of equation (\ref{eqn:inequality-1}). 
From equations (\ref{eqn:Jensen_for_theorem}) and (\ref{eqn:inequality-1}), 
\begin{align*}
&\mathbb{E}_{T}[F_{\mrm{AIS}}(S_T)] \nn
&\geq  -\ln Z_0 - \Big(\prod_{\mu=1}^N\sum_{\mbf{V}_{\mu}}\tau(\mbf{V}_{\mu})\Big) \ln \Big(\frac{1}{N}\sum_{\mu=1}^N \Lambda(\mbf{V}_{\mu})\Big)\nn
&= \mathbb{E}_{\tau}[F_{\mrm{mAIS}}(S_{\tau})]  \geq F
\end{align*}
is ensured; here, the final inequality is obtained from equation (\ref{eqn:F_mAIS_bound}). $\square$

\subsection*{Acknowledgment}
This work was partially supported by JSPS KAKENHI (grant numbers 18K11459, 18H03303, 20K23342, 21K11778, and 21K17804).

\bibliography{citation}
\end{document}